\documentclass[letterpaper, 10 pt, conference]{ieeeconf}  

\IEEEoverridecommandlockouts                              

\usepackage{multirow}
\usepackage{graphicx}
\usepackage{url,hyperref}
\usepackage{array}
\usepackage{xcolor,colortbl}
\usepackage[symbol]{footmisc}

\title{\LARGE \bf
Grasp Pre-shape Selection by Synthetic Training: Eye-in-hand Shared Control on the Hannes Prosthesis
}

\author{Federico Vasile$^{1,3}$ Elisa Maiettini$^{1}$ Giulia Pasquale$^{1}$ Astrid Florio$^{2}$ Nicolò Boccardo$^{2}$ Lorenzo Natale$^{1}$
\thanks{$^{1}$ Humanoid Sensing and Perception, Istituto Italiano di Tecnologia, Genoa, Italy (email: {\tt\footnotesize name.surname@iit.it})}%
\thanks{$^{2}$ Rehab Technologies Lab, Istituto Italiano di Tecnologia, Genoa, Italy (email: {\tt\footnotesize name.surname@iit.it})}%
\thanks{$^{3}$ DIBRIS, University of Genoa, Genoa, Italy}
}

\usepackage{eso-pic} 
\newcommand{\firstpagecopyright}{%
  \AddToShipoutPictureFG*{%
    \AtPageUpperLeft{%
      \hspace*{\dimexpr1in+\oddsidemargin\relax}%
      \raisebox{-3.5\baselineskip}[0pt][0pt]{%
        \begin{minipage}{\textwidth}
          \centering\footnotesize
          \textit{\textcopyright~2022 IEEE. Personal use of this material is permitted.
          Permission from IEEE must be obtained for all other uses, in any current or future media,
          including reprinting/republishing this material for advertising or promotional purposes,
          creating new collective works, for resale or redistribution to servers or lists, or reuse of
          any copyrighted component of this work in other works.}\\
          Preprint version (Sep.\ 2022). This work has been accepted for publication in IROS 2022.
        \end{minipage}%
      }%
    }%
  }%
}

\begin{document}
\firstpagecopyright   
\maketitle
\thispagestyle{empty}
\pagestyle{empty}

\begin{abstract}
We consider the task of object grasping with a prosthetic hand capable of multiple grasp types. In this setting, communicating the intended grasp type often requires a high user cognitive load which can be reduced adopting shared autonomy frameworks. Among these, so-called \textit{eye-in-hand} systems
automatically control the hand pre-shaping before the grasp, based on visual input coming from a camera on the wrist. In this paper, we present an \textit{eye-in-hand} learning-based approach for hand pre-shape classification from RGB sequences. Differently from previous work, we design the system to support the possibility to grasp each considered object part with a different grasp type. In order to overcome the lack of data of this kind and reduce the need for tedious data collection sessions for training the system, we devise a pipeline for rendering synthetic visual sequences of hand trajectories.
We develop a sensorized setup to acquire real human grasping sequences for benchmarking and show that, compared on practical use cases, models trained with our synthetic dataset achieve better generalization performance than models trained on real data. We finally integrate our model on the Hannes prosthetic hand and show its practical effectiveness. We make publicly available the code and dataset to reproduce the presented results\footnote[2]{\url{https://github.com/hsp-iit/prosthetic-grasping-simulation}}.

\end{abstract}

\section{Introduction}
\label{sec:introduction}
Latest advancements in the development of prosthetic arms~\cite{catalano2014,weiner2018} and specifically of myoelectric devices~\cite{patel2018,laffranchi2020} have led to the design of novel control systems based on electromyography (EMG) or mechanomyography (MMG) inputs. While effective, these systems need a significant cognitive effort for their control~\cite{gardner2020}. In fact, they generally require the user to execute specific muscle activation in order to drive the device. This increases the effort and the cognitive load required to get acquaintance with the prosthesis, contributing to the rejection of the device~\cite{simao2019}. 
Therefore, while representing useful information on user motor intentions, these sensors are not sufficient to achieve a simple and intuitive control of devices with several degrees of freedom. 
In this perspective, a \textit{shared autonomy} (or \textit{shared control}) of the device has been introduced~\cite{gardner2020}. The main idea is to split the task between the user and an automatic controller which relies on external contextual measurements as input. 
It has been shown in~\cite{mouchoux2021} that a semi-autonomous control of a prosthesis allows a user to accomplish tasks in shorter time and providing less explicit inputs with respect to volitional control only. In particular, for tasks like object grasping, in order to be useful for controlling the action, the sensing might happen remotely, with no physical contact (\textit{teleceptive sensing}~\cite{krausz2019}), before the actual action.
A recently explored solution is the integration of vision sensors (e.g., RGB and Depth cameras) to exploit visual information of the environment and of the other interacting elements. 
Specifically, for the grasping task in a \textit{shared autonomy}, the prosthesis control system could infer fundamental information from contextual measurements, such as (i) the user's target object, (ii) the part of the object that is intended to be grabbed and (iii) the intended use of the object. All these aspects concur to identify the grasp planned by the user (i.e., the \textit{grasp type}), which determines the correct hand pre-shaping (i.e., the \textit{pre-shape} phase), wrist orientation and hand closure, that would be otherwise unknown to the control system. Interestingly, it has been shown that the inclusion of visual information significantly increases the average grasp type classification accuracy~\cite{cognolato2022}. 
Specifically, in \textit{shared autonomy}, the hand pre-shaping and wrist orientation can be controlled automatically, based on visual input, while the closure of the fingers can be left to the user.\\
\begin{figure}
	\centering
	\includegraphics[width=0.9\linewidth]{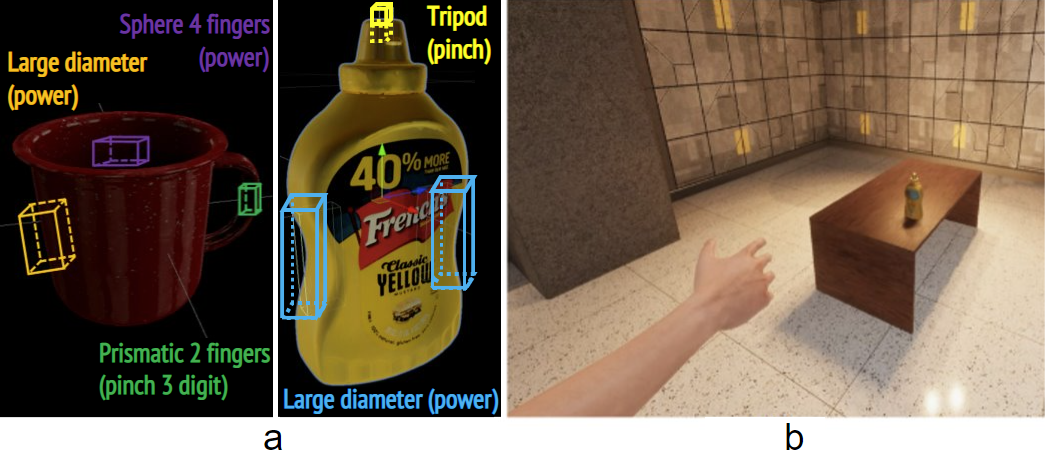}
	\caption{Examples of the proposed multi grasp type per object annotation for the Mug and the Mustard (each object part is labeled with a different grasp type and pre-shape) (a) and of the table-top setup rendering (b).}
	\label{fig:synthetic_data}
	\vspace{-0.55cm}
\end{figure}
\begin{table*}[h!]
	\vspace{2.5mm}
    \resizebox{17cm}{!}{
    \centering
	\begin{tabular}{c|c|c|c|c|c|c|c|c|c}
		\cline{1-10}
		\shortstack{\textbf{Pre-shape}\\ { } }   & \shortstack{Lateral\\ \textbf{ } } & \multicolumn{5}{c|}{ \shortstack{Power\\ \textbf{ } }} & \multicolumn{1}{c|}{\shortstack{Pinch\\ \textbf{ } }} & \multicolumn{2}{c}{\shortstack{Pinch\\ 3 Digit}}
		\\ \hline
		\shortstack{\textbf{ } \\  \textbf{Definition}\\ \textbf{ } }  & \shortstack{Adducted thumb\\ \textbf{ }}  & \multicolumn{5}{c|}{\shortstack{Central thumb\\ \textbf{ }} }   & \multicolumn{1}{c|}{\shortstack{Abducted thumb\\ \textbf{ }} }  & \multicolumn{2}{c}{\shortstack{Abducted thumb\\ \textbf{ }}}
		\\ \hline
		\shortstack{\textbf{Grasp type}\\ \textbf{ } }  & \shortstack{Adducted thumb\\ \textbf{ } }  & \shortstack{Large\\ diameter} & \shortstack{Small\\ diameter}& \shortstack{Medium\\ wrap} &\shortstack{Sphere 4\\ fingers} & \shortstack{Power \\sphere}  & \shortstack{Prismatic \\4 fingers} & \shortstack{Tripod\\ \textbf{ }}  & \shortstack{Prismatic\\ 2 fingers}  
		\\ \hline
        \shortstack{\textbf{Objects}\\ \textbf{ } \\ \textbf{ } \\ \textbf{ } } & \shortstack{Pitcher$^{1}$, Plate$^{2}$,\\ Spatula$^{2}$, Scissors$^{1}$ \\  \textbf{ }} & \shortstack{Chips can$^{2}$,\\ Mug$^{1}$,\\ Mustard$^{2}$}  & \shortstack{Hammer$^{1}$\\ \textbf{ } \\ \textbf{ } \\ \textbf{ } }  & \shortstack{Meat can$^{2}$\\ \textbf{ } \\ \textbf{ } \\ \textbf{ } } & \shortstack{Chips can$^{1}$,\\ Meat can$^{2}$,\\ Mug$^{2}$} & \shortstack{Plum$^{1}$, \\Baseball ball$^{1}$\\ \textbf{ }} & \shortstack{Spoon$^{1}$, Large marker$^{1}$, \\Scissors$^{1}$, Spatula$^{1}$, \\ Banana$^{1}$, Pitcher$^{1}$} & \shortstack{Red wood block$^{1}$,\\ Mustard$^{1}$,\\ Banana$^{1}$} & \shortstack{Mug$^{1}$\\ \textbf{ } \\ \textbf{ } \\ \textbf{ } }
        \\ \hline
	\end{tabular}
    }
	\caption{Association of pre-shapes with different grasp types and objects according to~\cite{wake2021,feix2015}, defining them with the thumb pose. The superscript digit on the object name refers to the number of object parts having that grasp type.}
	\label{table:grasp_type}
    \vspace{-0.9cm}
\end{table*}
In this work, we focus on the problem of the automatic control, with vision, of the hand pre-shaping before the grasp, hence addressing a so-called \textit{hand pre-shape classification} problem. Machine learning has been adopted in the recent literature for this task~\cite{ghazaei2017,taverne2019,gardner2020}. While effective, these methods require large and varied training sets. In the referenced literature on vision-based prosthetic control, the data is gathered for the purpose, by setting up data acquisition sessions that are typically long and tedious. In fact, while training deep learning systems on synthetic datasets is a quite established practice in computer vision and robotics, this approach is seldom adopted in prosthetics.
We present a methodology and related tool for the synthetic generation of human-like grasping RGB sequences, which allows to produce low-cost training sets for learning-based pre-shape classification. 
Remarkably, we show that models trained with the proposed synthetic dataset achieve comparable or higher performance on real use cases, than model trained on real data. We then present a prosthesis control pipeline, trained on the presented synthetic dataset, that allows to predict the user’s intended hand pre-shape and consequently move the prosthesis fingers accordingly.
Note that, differently from similar work~\cite{taverne2019}, we consider the case in which different object parts can be associated to different grasp types (and thus, pre-shapes). This is a critical aspect since it has been shown that one reason for amputees dissatisfaction with the prosthesis is the lack of the device adaptability to different object properties~\cite{kyberd2007}. Thus, the target is to enable the user to grasp an object from each side and for all possible usages.
Moreover, we develop a sensorized setup. We adopt it to empirically validate our design choices (e.g., the sensor placement). Then, we use it both (i) for collecting real human grasping sequences for benchmarking and (ii) for testing online the proposed approach. Finally, we integrate it with the Hannes prosthesis~\cite{laffranchi2020} to test the presented pipeline, showing its effectiveness.

This paper is organized as follows. In Sec.~\ref{sec:related_work}, we review the related literature. Then, in Sec.~\ref{sec:methods} and~\ref{sec:synthetic_dataset}, we describe the proposed approach for pre-shape classification and for human-like trajectories generation. In Sec.~\ref{sec:experiments}, we illustrate the experimental analysis carried out on datasets and with the Hannes prosthesis. Finally, in Sec.~\ref{sec:conclusions} we draw conclusions.

\section{Related Work}
\label{sec:related_work}
%
\noindent\textbf{Vision-aided prosthetic grasping}.
Recent methods that exploit visual cues for prosthetic grasping differ mainly for sensor placements. For instance,~\cite{chunyuan2020} uses an external static camera to acquire RGB and Depth images and to segment the target object via background subtraction. This information is then used by a learning-based pipeline to predict the corresponding grasp type. This is not suitable for real-world applications since it limits the work space to a static camera field of view and requires a preliminary sensor calibration. Moreover, it has been shown in~\cite{lampe2013} that static external cameras allow for a less accurate control with respect to other placements, like, on the prosthesis (\textit{eye-in-hand}~\cite{lampe2013}) or on a user's headset (\textit{egocentric} point of view).\\
In approaches with an \textit{egocentric} placement, the sensor has a wide field of view and additional information, as the user's gaze, is available. For instance,~\cite{markovic2014} and~\cite{markovic2015} use a stereo vision camera mounted on augmented reality glasses. The target object is identified when the user directs their gaze, and therefore the glasses, towards it. Then the geometrical properties of the object are retrieved by analyzing the Depth information from the camera and the system estimates the grasp type. 
In~\cite{zandigohar2021}, instead, a pre-trained object detector is fine-tuned for the purpose of grasp detection. The resulting grasp detector provides bounding boxes of possible objects to be grasped and the closest box to the user’s gaze is selected. This approach requires the user to direct gaze to the object of interest, and to wear a headset or glasses.
Conversely, the \textit{eye-in-hand} configuration can be completely transparent to the user and allows to gather closer views of the objects to grasp. This makes it easier to identify the target with visual and motion cues~\cite{he2020}. For instance, in~\cite{madusanka2017} the geometrical information (centroid and major axes) of the target object is inferred by using an RGB-D camera placed on the prosthesis and it is used to control the wrist orientation through visual servoing. In~\cite{dovsen2010}, instead, the same information is used to select the correct grasp type according to an \textit{IF-ELSE} set of rules. More recent approaches rely on deep learning techniques to extract useful information for the grasp like, target object segmentation~\cite{hundhausen2019,hundhausen2020,hundhausen2021}, object parts affordance~\cite{ragusa2021} and grasp type prediction~\cite{ghazaei2017,taverne2019}.

In this work, we compare the performance of learning-based methods in both configurations (\textit{egocentric} and \textit{eye-in-hand}) in the same setting. We find that the \textit{eye-in-hand} provides better visual cues for the considered task. Moreover, we develop a learning-based prosthesis control pipeline that allows to predict the user's intended hand pre-shape and consequently execute it on the device.

\noindent\textbf{Object grasping datasets}. To achieve good performance and generalization capabilities, supervised learning algorithms require quality annotated and abundant training data. For instance, recent work~\cite{saudabayev2018,chao2021,franziska2021} presents novel multi-modal datasets of humans grasping or manipulating objects which are remarkably valuable for benchmarking and performance analysis purposes. However, they are acquired with highly sensorized, extremely controlled setups. This approach does not scale to the need of a high variability in the data for training purposes. 
A recent trend for object grasping is to synthetically generate the data for model training. For instance, ~\cite{lin2020} generates a dataset under the hypothesis that common household objects can be decomposed into one or more primitive shapes each of those having a family of associated grasps. Other approaches in robotics, like~\cite{zhu2021}, identify contact maps on objects of interest, specifying the dexterous hand configuration for grasping it. However these approaches limit their contribution to generating the correct fingers position on the object, while in the task of pre-shape classification the arm trajectory towards the target is also important to discriminate between different pre-shapes for the same object. 
In prosthetics, the HandCam dataset~\cite{taverne2019} has been recently presented. The proposed setting is similar to the one used in this work. However, the dataset has a great limitation, i.e., it does not consider the realistic case in which different grasps can be associated to different parts of the same object~\cite{feix2014} and this reduces the problem to an object categorization. Finally, regarding synthetic visual data generation, this technique is rarely applied to prosthetics. To the best of our knowledge, only one work~\cite{zhong2020} has been proposed that generates a synthetic dataset for the task of target object identification.

In this work, we generate a synthetic dataset to address the task of hand pre-shape classification. Specifically, we consider the possibility to associate different grasp types to different object parts (see Fig.~\ref{fig:synthetic_data}a). This makes the task more difficult since identifying the object is not enough for correctly predicting the pre-shape. The arm trajectory towards the object part, during the approach, has to be considered as well. Finally, we apply domain randomization~\cite{tobin2017} to improve generalization capabilities of the model.

\section{Methods}
\label{sec:methods}
In the \textit{shared autonomy} framework, the prosthesis user and the control system need to cooperate in order to achieve the target grasping task. Ideally, this cooperation should be transparent to the user to allow a natural prosthesis utilization with the lightest cognitive load possible. At the same time, the user should feel to be in full control of the prosthesis and not vice-versa. To this aim, the start of a grasp movement, the timing for the fingers closure around the object and the force to apply could be left to the user (e.g., by means of muscle activation read through EMG sensors). On the contrary, the selection of the hand pre-shape can be delegated to the prosthesis control system. 
\begin{figure}
	\vspace{2.5mm}
	\centering
	\includegraphics[width=0.85\linewidth]{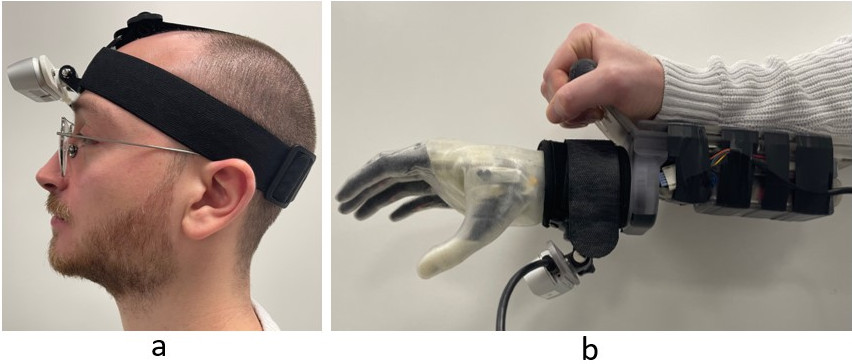}
	\caption{Pictures of the developed wearable sensorized experimental setup and of the used Hannes prosthesis~\cite{laffranchi2020}.}
	\label{fig:mockup}
	\vspace{-0.4cm}
\end{figure}

Thus, differently from previous work~\cite{taverne2019}, we focus on pre-shape (instead of grasp type) classification. We propose a system that can predict the best prosthesis pre-shape to grasp a given object, based on visual input. In order to exploit motion cues, we consider sequences of RGB frames taken in the time interval between the start of the arm movement towards the object and the moment before the hand touches the target. Note that, in this setting, the sequence starts when the user has their hand pointing to the object (i.e., in view from the \textit{eye-in-hand} camera) and finishes when the hand touches it for grasping it. Hence, we frame the task as a pre-shape classification of visual sequences. We consider four pre-shape classes, namely, \textit{Power}, \textit{Lateral}, \textit{Pinch} and \textit{Pinch 3 Digit} (refer to Tab.~\ref{table:grasp_type} for the thumb configuration for each pre-shape). Then, we add the class \textit{No grasp}, to discriminate when no pre-shape needs to be executed.
Finally, note that we hypothesize to have only one object in the scene. The extension to multiple objects would include a preliminary step of target object identification such as the ones proposed in~\cite{zhong2020,he2020} which is out of the scope of this paper but can be considered as future work.

In this paper we evaluate the advantages of using synthetic data for pre-shape classification learning. To this aim, we rely on two well-established learning-based models that share the same high-level architecture, i.e. (i) Convolutional Neural Network (CNN) based feature extraction (CNN backbone), followed by (ii) a pre-shape classifier. The first block encodes each image into a convolutional feature vector which is then taken as input by the classifier to predict a pre-shape for each image. In this work, we use \textit{Mobilnet V2}~\cite{sandler2018} as CNN backbone, pre-trained\footnote{\url{https://download.pytorch.org/models/mobilenet_v2-b0353104.pth}} on the ImageNet dataset~\cite{imagenet}. For pre-shape classification, we adopt either  \textit{Fully-connected layer} (\textit{CNN + FC}) or \textit{Long short-term memory}~\cite{hochreiter1997} (\textit{CNN + LSTM}). For the former, we use one FC layer of dimension \textit{C}, where \textit{C} is the number of classes. For the latter, we use an LSTM with 256 hidden units, followed by a FC layer with dimension \textit{C} to obtain the class scores. The input image size is 224x224px. The batch size is 32 when training on real dataset and 256 when training on synthetic dataset. In both cases, we fix the weights of the CNN backbone, while we train from scratch the classifier on top with a cross-entropy loss. In order to choose the number of training epochs, we evaluate performance on a validation set. Specifically, we reduce the learning rate (i.e. learning rate reduction on plateau) by a factor of 0.1 if the validation loss does not decrease after $6$ epochs, and we stop the training (i.e. early stopping) if the validation loss does not decrease after $10$ epochs (we initialize the learning rate to $0.0005$). Moreover, in order to counteract any unbalancing in the dataset, before starting each epoch we randomly downsample every class to the cardinality of the minority one.

\label{sec:methods:preshapepred}
Finally, the models are used to predict a pre-shape for each sequence. Specifically, they predict one of the four pre-shapes (or the class \textit{No grasp}) for each frame. The final prediction for the sequence is given by a majority voting, i.e., the pre-shape class that has been predicted for most of the frames (excluding the class \textit{No grasp}).

\section{Synthetic Data Generation}
\label{sec:synthetic_dataset}
\begin{figure*}
	\vspace{2.5mm}
	\centering
	\includegraphics[width=0.85\linewidth]{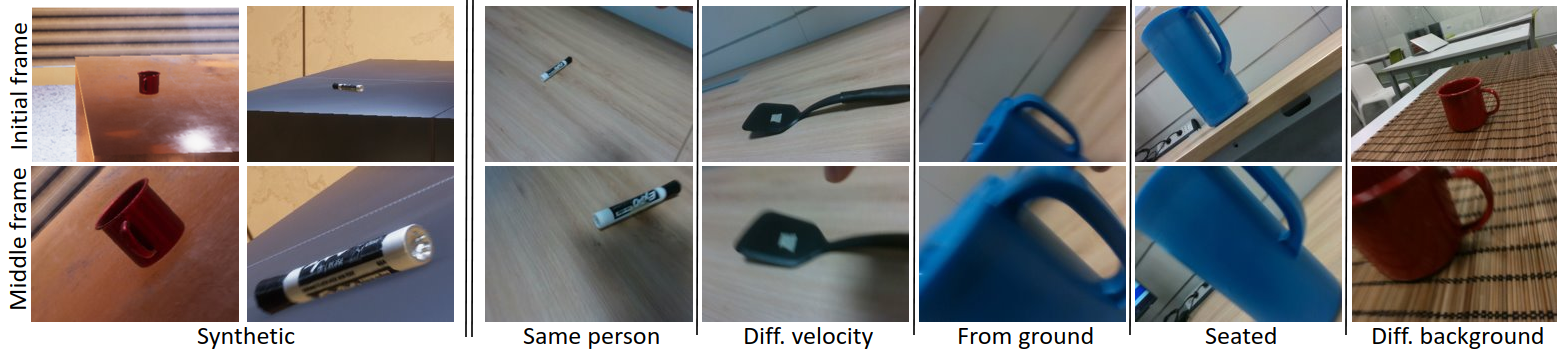}
	\caption{Sample frames of the proposed synthetic dataset and of the collected real sequences for the different training and test sets.}
	\label{fig:sample_frames}
	\vspace{-0.5cm}
\end{figure*}

We simulate the work space as a room with a table in the center (refer to Fig.~\ref{fig:synthetic_data}b, for an example of the simulated setting). We render a human arm and we rigidly attach a RGB sensor below the wrist, simulating the view a camera would have if placed as in Fig.~\ref{fig:mockup}b. Then, we place the 3D model of an object on the table and we simulate a human arm trajectory that makes the hand reach the object to grasp.

\noindent\textbf{Different grasp types for different object parts.}
For this work, we chose a subset of 15 objects from the YCB-Video~\cite{xiang2018} dataset (refer to the last row of Tab.~\ref{table:grasp_type} for the list of all the objects) and the corresponding 3D models\footnote{\url{http://ycb-benchmarks.s3-website-us-east-1.amazonaws.com}}. We choose objects from this dataset in order to ease experiments reproducibility and comparison with the literature. However, the same approach can be applied to any object with a 3D model available. Firstly, we assign one grasp type to each graspable object part. Specifically, we take inspiration by~\cite{wake2021} for the objects grasp types association but we exclude or modify those that are not possible to execute with our prosthesis. Then, since in the \textit{shared autonomy} framework we aim to address a pre-shape classification task, we associate a pre-shape to each considered grasp type, following~\cite{feix2015}. Tab.~\ref{table:grasp_type} overviews the resulting associations of the chosen grasp types, pre-shapes and objects.
Note that, in this way, we create a link between each object part and the corresponding grasp type and pre-shape. In order to implement this in the rendering pipeline, we overlay a parallelepiped to each object part of interest and we assign it the correct grasp type (refer to Fig.~\ref{fig:synthetic_data}a for grasp type annotation examples). If, while executing an arm trajectory, the hand clashes with one parallepiped, the frame sequence is annotated as the corresponding grasp-type.
Note that, differently from previous work~\cite{taverne2019}, we consider the case in which different object parts can be associated to different grasp types. This is a fundamental aspect because, in practice, one object can be grasped differently from different parts, depending on its use~\cite{feix2014}. In this case, simply recognizing the object is insufficient, we show in Sec.~\ref{sec:experiments} that such a na\"ive approach brings to a grasping performance degradation even in the ideal case of 100\% object classification accuracy.

\noindent\textbf{Human-like arm trajectory generation.}
The arm trajectory towards the object part, together with the object type, conveys useful information to predict the grasp type. Hence, simulating realistic trajectories is key in this setting. To this purpose, we model the arm and the hand as a mesh with a RGB camera rigidly attached below the wrist. In this setting, the trajectory to model consists of the movement of a 6D point in space. Since our input is only RGB frames, we aim to model a camera trajectory resembling the one that the camera would have if mounted on a real person approaching the target object with their hand, to grasp it. In our setup (both real and synthetic), we placed the camera such that no finger occlusions occur. However, since we adopt a photorealistic arm model\footnote{\url{https://www.cgtrader.com/free-3d-models/character/man/fps-arms-pack}}, simulating different camera placements ---including such occlusions--- would be possible if needed.
To make a realistic hand approach, referring to the literature on human arm movements~\cite{flash1985}, we implement a minimum jerk trajectory, i.e., on a straight line, with a bell-shaped velocity profile. Moreover, to simulate the pronation-supination and flexion-extension of the wrist at the end of the reaching, we rotate the camera. In particular, while approaching the object, the hand rotates such that at the end of the movement the normal to the palm becomes parallel to 
the normal to the external face of the parallelepiped linked to the target object part.

\noindent\textbf{Domain randomization.}
In order to cover the well-known sim-to-real gap \cite{tobin2017} and to generalize to novel conditions, it is necessary to introduce variability in the synthetically generated training set. To this end, by following a well-established practice in computer vision~\cite{tobin2017}, we randomize the textures of the room walls, floor and table and we randomly vary the light conditions. Moreover, we randomize the object pose on the table (refer to Fig.~\ref{fig:sample_frames}, for example of generated frames).
Finally, we addressed the problem of introducing sources of randomization in the process of arm trajectory generation. Differently from the above aspects, this one is peculiar of the considered application and hence we devised a novel randomization strategy. Specifically, for each simulated approaching trajectory we fix the final point as the center of the chosen parallelepiped, whereas we randomly select the initial one. Precisely, this is randomly sampled from a plane perpendicular to the table-top and placed at a distance from the table border such that both the object and part of the background are in view. Before a sequence execution, given an initial and final points couple, the following procedure is executed. Firstly, a straight line trajectory is generated to check for collisions. Then, if the collision with the considered parallelepiped happens, the trajectory is executed and the sequence is labeled with the grasp type represented by the parallelepiped. Otherwise, e.g., if the trajectory collides with the object mesh, it is discarded. This ensures that given a \textit{grasp type - object part} pair, only reasonable sequences are generated (e.g., it is impossible to grab a mug by the handle if this is not visible).
Thus, the final dataset is composed of video sequences taken from the RGB camera on the wrist of the simulated arm reaching the object. Each sequence is annotated with the chosen object, grasp type and pre-shape.

We developed the described data generation approach using the Unity\footnote{\url{https://unity.com/}} simulation engine to obtain photo-realistic frame sequences and integrating the Perception package\footnote{\url{https://github.com/Unity-Technologies/com.unity.perception}} to ease domain randomization and sequence labeling.



\section{Experiments}
\label{sec:experiments}

\subsection{Experimental setup}
\label{sec:experiments:setup}
    In order to evaluate the proposed synthetic data generation technique, we generate a dataset of 47 sequences for each \textit{grasp type - object part} pair (as defined in Tab.~\ref{table:grasp_type}), resulting into 1457 videos. Notice that the superscripts on the object names in Tab.~\ref{table:grasp_type} indicate the number of object parts having the same grasp type (e.g., the mustard in Fig.~\ref{fig:synthetic_data}a has the \textit{large diameter} grasp type on both bottle sides). Where not differently specified, we use this dataset to train the considered learning models.
    
    With the aim to test all the components of the proposed work, we develop a wearable sensorized setup composed of two Intel RealSense D435 cameras: one is worn on the head and the other on the wrist, with two elastic bands. We use this setup both to collect sequences for benchmarking purposes and to test the control pipeline directly on the Hannes prosthesis. Specifically, three healthy subjects collected several sequences. The first subject stands in front of the object to grab, placed on a table-top and collected 311 sequences. For each \textit{grasp type - object part} pair, he was asked to vary the approaching direction and object pose on the table. Each sequence is 3 seconds long, where the first 2.5 seconds are labeled as the grasp type executed by the subject and the last 0.5 seconds as \textit{No Grasp} (since the corresponding frames show the moment of hand-object contact). We split this set of sequences into three sub-sets, balancing the different objects and grasp types presence. This results in sub-sets of 46, 58 and 207 sequences. We use the first as a validation set for model training to choose the number of optimization epochs as described in Sec.~\ref{sec:methods}. The second one, referred to as \textit{Same person}, is used as a test set to evaluate the performance of the proposed approach for this subject in the following sections. Finally, the third one is used as a real training set, for comparing the performance of models trained on it with the proposed synthetic dataset. Refer to Fig.~\ref{fig:sample_frames} (\textit{Same person}) for example frames of this dataset.
    Then, other two healthy subjects collected four further sets, performing two trials for each \textit{grasp type - object part} in each set. Refer to Fig.~\ref{fig:sample_frames} for example frames of each of them. The sets are designed as follows:
    \begin{itemize}
        \item \textit{Different velocity}. The grasp is completed in 1.5 seconds instead of 3 seconds. The hand movement starts from the same distance to the object as for the first subject, thus resulting in a faster approach.
        \item \textit{From ground}. Rather than starting with the object in view, (i) the arm is initially extended along the side, (ii) the subject raises their arm, (iii) approaches the object and (iv) grasps it. All steps are performed smoothly. Starting with the arm along the side produces different trajectories resulting in different object views. For our experiments, the first part of the sequence is trimmed out at inference time as the camera points at the ground.
        \item \textit{Seated}. The subject performs the grasp while seating in front of the table. The different body posture influences the initial hand pose with respect to the object and thus the approach. The grasp is completed in 1.5 seconds.
        \item \textit{Different background}. We vary the tablecloth underneath the object to be grasped. We do this to evaluate the system performance on a different background.
    \end{itemize}
Since the considered learning models output a prediction for each frame but the final prediction for the entire sequence is given by majority voting (see Sec.~\ref{sec:methods}), the performance is evaluated by considering a per-video Accuracy.
Results are reported as mean and standard deviation over 3 trials of the same experiments, by varying the random seed.

\subsection{System design motivation}
\label{sec:experiments:motivation}
\begin{figure}
	\centering
	\includegraphics[width=0.75\linewidth]{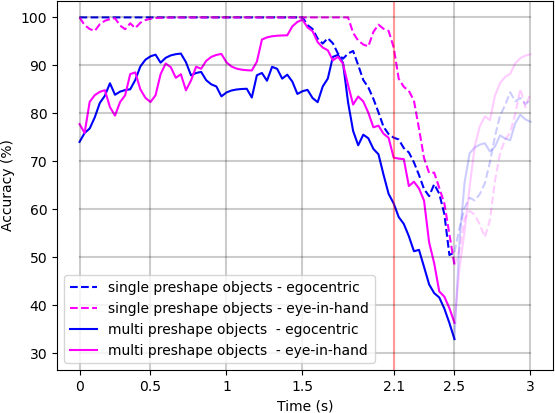}
	\caption{Accuracy trends comparison over time during a grasping sequence of models trained with egocentric (blue) and eye-in-hand (pink) data. We report performance for single (dashed) and multi grasp (solid) objects.}
	\label{fig:time_plot}
	\vspace{-0.3cm}
\end{figure}

In this section, we empirically motivate our main design choices.\\
\noindent{\textbf{One object, multiple grasps}}.
In order to demonstrate the importance of having different grasps for different object parts, we show the performance loss that one would achieve on our dataset with a model predicting only one grasp type for each object. For doing this, we consider an ideal model that can discriminate with 100\% accuracy among the 15 objects chosen for our experiments. Then, an object-grasp type mapping is required. Hence, we consider the grasp type-object association of Tab.~\ref{table:grasp_type}. Note that, for objects with multiple grasp possibilities, the most frequent one is chosen.
This theoretical model is evaluated on the \textit{Same person} test set (see the first row of Tab.~\ref{table:setting_validation}) that we labeled with multiple grasps per object. By using the theoretical single grasp model, a user could correctly grasp the object only 79.3\% of the time. Precisely, single grasp objects are predicted with 100\% accuracy, while multi grasp objects obtain only 56\% accuracy even considering perfect object classification. This motivates the need for considering different grasp types for different object parts when learning the pre-shape classification model and confirms that our dataset well reflects this challenge. Moreover, the higher accuracy obtained by our model (see second row in Tab.~\ref{table:setting_validation}) confirms its disambiguation capability between different pre-shapes for the same object. \\
\noindent{\textbf{Egocentric vs eye-in-hand}}. 
The camera positioning is another crucial aspect of the system design. It affects the quality of the collected RGB sequences and therefore the prediction performance. Moreover, some placements might be invasive and affect the prosthesis usability. To compare the two configurations, we consider data acquired by the first subject with our setup (see Sec.~\ref{sec:experiments:setup}) from both cameras (on head and hand) and we use them to train and test the two different models presented in Sec.~\ref{sec:methods}. Results are presented in Tab.~\ref{table:setting_validation}. In general, eye-in-hand models obtain higher performance and less variability than the egocentric ones. Additionally, in Fig.~\ref{fig:time_plot} we report the per-video accuracy achieved at each frame by the two CNN + FC models throughout the approaching sequence. We compare performance obtained for single grasp (dashed line) and multi grasp (solid line) objects. 
As it can be observed, regarding single grasp objects, in both setups the model identifies the correct pre-shape since the beginning of the sequence (because in this case this is uniquely identified by the recognized object). Differently, for multi grasp objects, the eye-in-hand data clearly brings more information to discriminate between different pre-shapes for the same object between second $0.7$ and $1.7$. This may be due to the fact that as the camera gets closer to the object, the target part to grasp becomes the most visible one. This is shown by the accuracy peak at time 1.5s. 
Note that, there is a drop for all models around time 2.5s. This is due to the label switch from pre-shape to \textit{No grasp}. 
To further support our design choice, we notice that past literature~\cite{jeannerod1984} shows that the maximum hand aperture, during grasping, happens at 70\% of the reaching phase (i.e., $\sim$2.1s in our case). Since for our setting and for the considered sequences length, the performance peak for the eye-in-hand configuration happens at $\sim$1.5s, we believe that this is a promising result to obtain a smooth reach-and-grasp movement since the prediction seems to come in useful time to actuate the hand aperture.
Considering also the minor invasiveness of the eye-in-hand configuration, we consider this configuration for our work.\\
\noindent{\textbf{FC vs LSTM}}. Previous work~\cite{taverne2019} suggests that recurrent models (CNN + LSTM) should outperform single frames ones (CNN + FC) for the task of hand pre-shape classification. However, our problem is different from~\cite{taverne2019} since we consider the case in which the same object can be associated to different grasp types. Therefore, we compare the two different learning architectures described in Sec.~\ref{sec:methods} in our multi grasp setting. To this aim, we consider the data acquired by the first subject and we use it to train and test the two architectures. Results are presented in Tab.~\ref{table:setting_validation}. As it can be observed, in our case, the single frames model (CNN + FC) clearly outperforms the recurrent one (CNN + LSTM). Interestingly, all the errors for the recurrent models come from the multi grasp objects. This is due to the fact that, as previously discussed, the first frames of the grasping sequences are ambiguous for those objects. This ambiguity is an issue especially for the recurrent models since for them it is more difficult to recover from an early error than for the single frames one. Note that, considering only single grasp objects as in~\cite{taverne2019} might hide this issue. Given these results, we consider the CNN + FC architecture for our work.
\begin{table}[]
	\vspace{2.5mm}
	\centering
	\resizebox{8.5cm}{!}{
	\begin{tabular}{c|c|c|c}
		\cline{1-4}
		\multicolumn{2}{c|}{\textbf{ }} & \textbf{\shortstack{Egocentric \\ Video acc (\%))} }   & \textbf{\shortstack{Eye-in-hand \\ Video acc (\%))}}                           \\ \hline
		\shortstack{Single-grasp\\ Training}                & \multicolumn{1}{c|}{\shortstack{Perfect \\ classification}}   & \shortstack{$79.3 \pm 0.0$\\\textbf{ }} 
		                                                                                                                    & \shortstack{$79.3 \pm 0.0$\\\textbf{ }} \\ \hline
		\multirow{2}*{\shortstack{Multi-grasp\\ Training}}  & \multicolumn{1}{c|}{CNN + FC}                                 & $96.6 \pm 1.4$  & $98.9\pm 0.8$         \\
		                                                    & \multicolumn{1}{c|}{CNN + LSTM}                                & $87.4 \pm 3.5$  & $92.0 \pm 2.1$        \\ \hline
	\end{tabular}
	}
	\caption{Comparison between models trained with the egocentric and eye-in-hand configurations for both  single grasp (case of perfect classification) and multi grasps per object settings.}
	\label{table:setting_validation}
	\vspace{-0.3cm}
\end{table}
\subsection{Learning pre-shape classification from synthetic data}
One major contribution of this work is the development of a synthetic visual data generation tool for prosthetic grasping. In this section, we evaluate the effectiveness of the proposed approach by using the generated data to train a learning-based model for pre-shape classification. We compare the obtained performance against a model trained on real data, i.e., on the 207 sequences collected with the first subject, by testing on different real use case conditions, namely, over the test sets collected with our sensorized setup as described in Sec.~\ref{sec:experiments:setup}. We report results in Tab.~\ref{table:synthetic_training}.

Firstly, we evaluate the models on the \textit{Same person} test set (first row in Tab.~\ref{table:synthetic_training}). In this case, the real training and test sets are drawn from the same distribution, being acquired by the same subject under the same visual conditions, while varying only the approach direction. Conversely, the synthetic training set has been generated with no knowledge of the subject posture and background. As expected, in this case a model trained on the real dataset fits extremely well the user and visual conditions while the one trained on synthetic data performs worse. Interestingly, while the overall accuracy of $80.2\%$ is close to the $79.3\%$ obtained by the theoretical model trained on single-grasp labels discussed previously, in this case, $38.1\%\pm 0.02$ of the wrong predictions come from single grasp objects while $61.9\%\pm 0.02$ come from multi grasp ones.


Then, we analyze the performance of the same models in case of trajectory variations of the arm during the approaching phase (rows from 2 to 4 in Tab.~\ref{table:setting_validation}). Notably, in the first two test sets the synthetic model achieves performance comparable to the real one, and in the third test set the synthetic model performs clearly better. Specifically, the \textit{Seated} set is the most challenging one since its sequences present a more curved trajectory and consequently different viewpoints while approaching the object. While the real model exhibits a more pronounced drop in this case, the synthetic model has a smaller decrease. This confirms the robustness of the trajectories generated in the proposed synthetic dataset.


Finally, we compare the accuracy obtained by the learned models on the \textit{Different background} set (last row in Tab.~\ref{table:synthetic_training}). The different pattern in the background produces a performance gap for both the real and synthetic models with respect to the accuracy on \textit{Same person}. However, this gap is remarkably smaller ($\sim 4\%$) for the synthetic model than for the real model ($\sim 42\%$). This confirms results of previous work on domain randomization~\cite{tobin2017} and demonstrates the effectiveness of the proposed approach for pre-shape classification under different visual conditions.

As a final remark, these experiments show that except from tests on the same subject used for training, models trained on the synthetic dataset have comparable performance to models trained on real data. Moreover, in case of more challenging conditions in terms of arm movements and background variations, the synthetic dataset resulted to be crucial, while the real one led to a significant performance drop. Therefore, a real training set can be considered the best solution only for scenarios in which a dataset is available and the conditions are known, as shown in a similar work~\cite{taverne2019}. However this is practically unfeasible for prosthetic applications. Moreover, we show that the proposed approach to model eye-in-hand camera trajectories is not only effective to fill the sim2real gap but also achieve higher trajectory generalization.
 
\begin{table}[]
	\vspace{2.5mm}
	\centering
	\begin{tabular}{c|c|c}
		\cline{1-3}
		  & \textbf{\shortstack{Real tr. set \\ Video acc (\%))} }   & \textbf{\shortstack{Synthetic tr. set \\ Video acc (\%))}}    \\ \hline
		\multirow{1}*{\shortstack{Same person}}                      & $98.9\pm 0.8$  & $80.2 \pm 0.9$  \\ \hline \hline
		\multirow{1}*{\shortstack{Different velocity}}   & $81.7\pm 0.9$  & $79.7 \pm 0.8$  \\ \hline
        \multirow{1}*{\shortstack{From ground}}                     & $76.2\pm 1.0$  & $76.0 \pm 0.9$  \\ \hline

        \multirow{1}*{\shortstack{Seated}}                       & $63.9 \pm 1.0$ & $68.1 \pm 1.0$  \\ \hline \hline
        \multirow{1}*{\shortstack{Different background}}          & $56.2 \pm 1.7$ & $76.4 \pm 2.0$  \\ \hline
	\end{tabular}
	\caption{Video accuracy of the models trained on a real and on the proposed synthetic datasets, compared on different use cases}
	\label{table:synthetic_training}
	\vspace{-0.8cm}
\end{table}

\subsection{Application on the Hannes prosthesis}
\label{sec:experiments:application}

As a further contribution, we test the presented eye-in-hand shared control system on an improved version of the Hannes prosthetic hand~\cite{laffranchi2020}. Hannes is a myoelectric poly-articulated prosthesis equipped with electromyographic sensors, a battery pack designed to last up to 1 day, and control electronics, all placed inside the wearable socket (refer to~\cite{laffranchi2020} for further details). Remarkably, for object grasping, the version of Hannes used in this work comprises an active abduction/adduction joint for the thumb and a three-digital modality can be enabled. This allows to execute the four different pre-shapes considered in this work (namely, \textit{Power}, \textit{Lateral}, \textit{Pinch} and  \textit{Pinch 3 Digit}). 

A more advanced eye-in-hand system prototype would have the RGB camera embedded into the palm of the hand prosthesis. However, for the first testing of our pipeline on Hannes, we integrate it with our sensorized setup and we use an external able-bodied adapter to allow a healthy user to drive it. The final setup is shown in Fig.~\ref{fig:mockup}. All the computation for pre-shape classification is performed on an external GPU-enabled laptop which is connected to the on board control electronics of Hannes via serial connection.
The experiment is designed as follows. A healthy user, handling the prosthesis, stands in front of the object to grasp placed on a table. After a starting trigger (for this analysis we considered a keyboard signal), the user approaches the object, moving the prosthesis towards it as to grasp it. The approach lasts $\sim$2 seconds and the system collects RGB frames from the wrist-mounted camera. These frames are fed to the classifier (the CNN + FC model trained on synthetic data from Sec.~\ref{sec:experiments}) and a pre-shape is predicted. The pre-shape class is converted into the corresponding thumb and three-digital configuration and this is sent to the on board control electronics which translates it into control signals for the joints. Next, the pre-shape is executed and then the fingers close around the object. Note that, the aim of this experiment is to integrate the proposed approach with a prosthesis and evaluate performance, therefore in order to automatise the starting trigger and the fingers closure we just temporised them, but we plan to implement more refined approaches also for these steps (e.g., with EMG signals). A video demonstration of the resulting system is attached as supplementary material and it shows the usability of the proposed approach for prosthetic grasping.



\section{Conclusions}
\label{sec:conclusions}
We present an eye-in-hand shared control pipeline for prosthetic grasping. Specifically, we tackle the problem of hand pre-shape classification of RGB sequences from a wrist mounted camera with a learning-based approach. Generalization capabilities to different conditions (e.g., backgrounds and arm trajectories) are critical for prosthetic applications. To address this, 
we develop a synthetic human-like trajectories generation tool and we demonstrate that models trained with the proposed synthetic dataset achieve comparable or higher performance than models trained on real data on practical use cases. To the best of our knowledge, this is the first attempt to use synthetic visual data for the task of pre-shape classification for prosthetic grasping. 

Differently to previous literature, we considered the case where different grasp types can be associated to different object parts. While increasing the task complexity, this aspect is fundamental in order to devise an effective prosthesis control system, and thus increase the user satisfaction. We show that, by using well established predictive models, the eye-in-hand configuration allows to get better results in this setting than the egocentric one. Classification performance might be improved by exploiting more sophisticated techniques, such as the ranking labels~\cite{zandigohar2019}. Furthermore, the current model and data generation technique can be extended to deal with a more fine-grained supervision. 
For instance, each object part can be segmented and associated to the corresponding pre-shape, leading to the affordace segmentation task~\cite{ragusa2021}.





\section*{Acknowledgment}

This work was supported in part by the Istituto Nazionale Assicurazione Infortuni sul Lavoro, under the project iHannes (PR19-PAS-P1).
\vspace{-0.15cm}
\bibliographystyle{IEEEtran}
\bibliography{ihannes_bibliography} 


\end{document}